\documentclass[letterpaper]{article} 
\usepackage{aaai25}  
\usepackage{times}  
\usepackage{helvet}  
\usepackage{courier}  
\usepackage[hyphens]{url}  
\usepackage{graphicx} 
\usepackage{natbib}  
\usepackage{caption} 
\usepackage{algorithm}
\usepackage{algorithmic}
\usepackage{amsmath}
\usepackage{amssymb}
\usepackage{bm}
\usepackage{booktabs}
\usepackage{pifont}
\usepackage{tcolorbox}
\newcommand{\cmark}{\textcolor{green}{\ding{51}}}
\newcommand{\xmark}{\textcolor{red}{\ding{55}}}

\usepackage{newfloat}
\usepackage{listings}
\DeclareCaptionStyle{ruled}{labelfont=normalfont,labelsep=colon,strut=off} 
\floatstyle{ruled}
\newfloat{listing}{tb}{lst}{}
\floatname{listing}{Listing}

\usepackage{fancyhdr}
\fancypagestyle{firstpagehf}
{
    \fancyhf{}
    \fancyhead[HC]{The AAAI 2025 Workshop on Advancing LLM-Based Multi-Agent Collaboration (WMAC)}
    \fancyfoot[C]{\thepage}
}
\nocopyright

\title{InvAgent: A Large Language Model based Multi-Agent System for\\Inventory Management in Supply Chains}
\author{
    Yinzhu Quan\equalcontrib,
    Zefang Liu\equalcontrib
}
\affiliations{
    Georgia Institute of Technology\\
    Atlanta, GA 30332, USA\\
    yquan9@gatech.edu, liuzefang@gatech.edu
}

\begin{document}
\thispagestyle{firstpagehf}
\maketitle

\begin{abstract}
Supply chain management (SCM) involves coordinating the flow of goods, information, and finances across various entities to deliver products efficiently. Effective inventory management is crucial in today's volatile and uncertain world. Previous research has demonstrated the superiority of heuristic methods and reinforcement learning applications in inventory management. However, the application of large language models (LLMs) as autonomous agents in multi-agent systems for inventory management remains underexplored. This study introduces a novel approach using LLMs to manage multi-agent inventory systems. Leveraging their zero-shot learning capabilities, our model, InvAgent, enhances resilience and improves efficiency across the supply chain network. Our contributions include utilizing LLMs for zero-shot learning to enable adaptive and informed decision-making without prior training, providing explainability and clarity through chain-of-thought, and demonstrating dynamic adaptability to varying demand scenarios while reducing costs and preventing stockouts. Extensive evaluations across different scenarios highlight the efficiency of our model in SCM.
\end{abstract}

\section{Introduction}

Supply chain management (SCM) involves coordinating and managing the flow of goods, information, and finances across various interconnected entities, from suppliers to consumers, to deliver products efficiently and effectively. Inventory management, a critical component of SCM, focuses on overseeing and controlling the ordering, storage, and use of components and finished products. In today's volatile, uncertain, complex, and ambiguous (VUCA) world, effective inventory management is essential for aligning supply with demand, minimizing costs, and enhancing the resilience of supply chains. This ensures that companies can adapt to disruptions \cite{quan2023predictive}, optimize resources \cite{abaku2024theoretical}, and maintain seamless operations \cite{yasmin2024supply} in a highly interconnected and dynamic market environment.

Previous research in inventory management has explored various applications of heuristic methods, such as the beer distribution game  \cite{goodwin1994beer,edali2014mathematical,oroojlooyjadid2022deep}. Additionally, numerous implementations of reinforcement learning models have been investigated, including the decentralized inventory management \cite{mousa2024analysis} and the adaptive supply chain synchronization \cite{kegenbekov2021adaptive}. However, these methods often require sophisticated design, extensive training resources, and lack explainability. In contrast, large language models (LLMs) present a promising alternative, offering adaptive decision-making without prior training and enhanced interpretability. Recent studies have started to utilize LLMs in supply chain research, as demonstrated by \citet{li2023largeb}, \citet{quan2024econlogicqa}, and \citet{singla2023empirical}.

\begin{figure*}[!h]
    \centering
    \includegraphics[width=.7\linewidth]{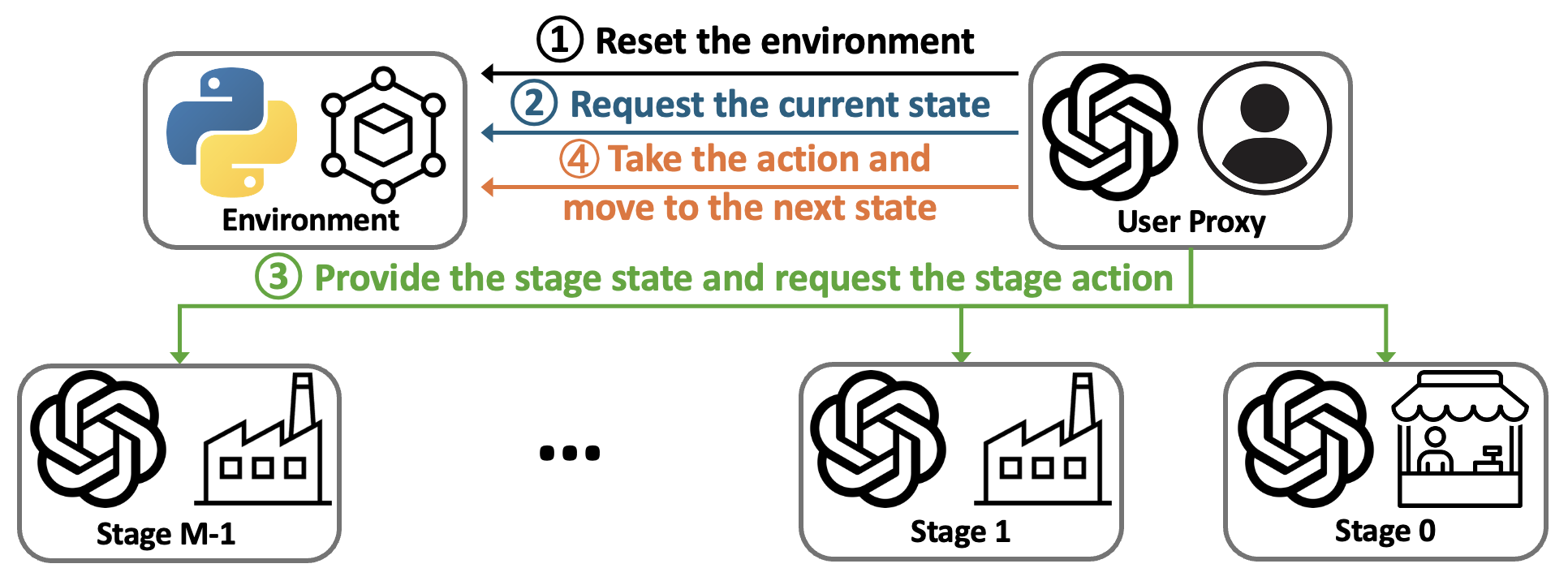}
    \caption{The framework of InvAgent, a LLM-based zero-shot multi-agent inventory management system. Firstly, the user proxy resets the environment at the beginning of the first round. Secondly, the user proxy requests the state of the current round for each stage from the environment. Then, the user proxy provides the current state to each stage and requests the action from it. Finally, all agents take actions together and move to the next state.}
    \label{fig:framework}
\end{figure*}

LLMs are increasingly utilized as autonomous agents in multi-agent systems, showcasing advanced planning, decision-making, and simulation capabilities across diverse domains \cite{guo2024large} such as gaming \cite{mao2023alympics} and financial markets \cite{li2023tradinggpt}. However, the application of LLMs to tackle the multi-agent inventory management problem (IMP) within supply chains remains relatively underexplored. In this study, we propose InvAgent, an advanced zero-shot multi-agent inventory management system utilizing LLMs. Our approach leverages LLMs to enhance system resilience and foster collaboration across various components of the supply chain network through their reasoning and decision-making capabilities. The framework of our paper could be seen from Figure \ref{fig:framework}.

Our contributions of this paper are as follows:
\begin{enumerate}
    \item We leverage LLMs to manage multi-agent inventory systems as zero-shot learners, enabling adaptive and informed decision-making without prior training or specific examples.
    \item Our model offers significant explainability and clarity, enhanced by chain-of-thought (CoT) for reasoning, making it easier to understand and trust, and more reliable compared to traditional heuristic and reinforcement learning models.
    \item Our model adapts dynamically to varying demand scenarios, minimizing costs and avoiding stockouts, demonstrating efficiency in supply chain management through extensive evaluation across different scenarios.
\end{enumerate}

\section{Related Work}

\textbf{LLM-Based Multi-Agent System Applications in Economics.} The LLM-based MASs have been used in economic and financial trading simulations to model human behavior. It enables agents with specific endowments, information, and preferences to interact in scenarios like macroeconomic activities \cite{li2023large}, information marketplaces \cite{weiss2023rethinking}, financial trading \cite{li2023tradinggpt,yu2024finmem}, and virtual town simulations \cite{zhao2023competeai}. These agents operate in both cooperative and decentralized environments, demonstrating diverse applications in economic studies \cite{guo2024large}.

\textbf{Multi-Agent System Applications in Supply Chain.} There are some studies extensively explore the potential of MAS to enhance supply chain efficiency and responsiveness, addressing various challenges from integration to dynamic adaptation and coordination. \citet{nissen2001agent} examines the integration of supply chains using agent-based technologies, highlighting how agents can facilitate more efficient and responsive supply chain operations. \citet{kaihara2003multi} discusses the application of MASs in modeling supply chains that operate in dynamic environments, focusing on how agents can adapt to changes and uncertainties. \citet{moyaux2003multi} explores how multi-agent coordination mechanisms can help reduce the bullwhip effect in supply chains, using a token-based approach to enhance collaboration and information sharing among agents.

\textbf{Multi-Agent Reinforcement Learning Applications in Supply Chain.} Research in multi-agent reinforcement learning (MARL) for SCM focuses on optimizing interactions and cooperation among multiple agents in dynamic environments. \citet{oroojlooyjadid2022deep} propose the Shaped-Reward Deep Q-Network (SRDQN) algorithm for RL in the beer distribution game, where agents optimize behaviors through rewards and punishments to improve performance. \citet{hori2023improving} enhance cooperative policies in the beer game using reward shaping techniques based on mechanism design applied to SRDQN, improving performance in multi-agent settings. Additionally, OR-Gym \cite{hubbs2020or} is an open-source library that benchmarks RL solutions against heuristic models in operations research problems including SCM.

\section{Methodology}

This section outlines the methodological framework, starting with the definition of an inventory system and then proposing our model InvAgent designed for supply chain optimization.

\subsection{Problem Definition}

\begin{figure}[h]
    \centering
    \includegraphics[width=.9\linewidth]{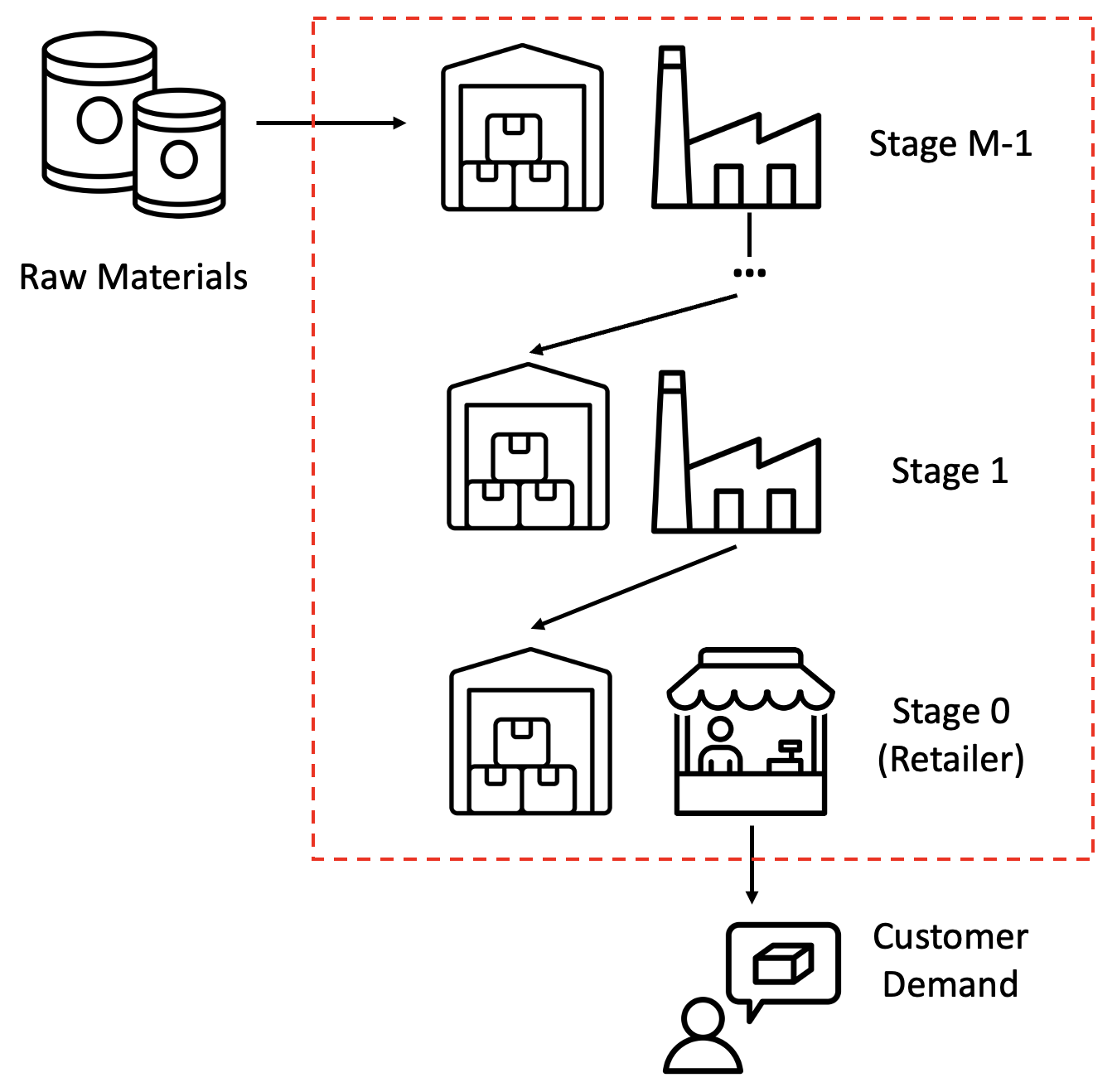}
    \caption{The flowchart of multi-echelon supply chain inventory management. Raw materials flow through each stage, comprising inventory storage and manufacturing facilities. The upstream factory at stage $i$ supplies intermediate products to the downstream stage $i-1$, where they are stored as inventory. Stage 0 (retailer) provides final products to satisfy customer demand.}
    \label{fig:flowchart}
\end{figure}

A multi-period, multi-echelon inventory system for a single non-perishable product is designed for illustrating and simulating a typical multi-stage supply chain. As shown in Figure \ref{fig:flowchart}, each stage in this supply chain consists of an inventory holding area and a production area. The inventory holding area stores the materials necessary for production at that stage. One unit of inventory produces one unit of product at each stage. There are lead times for transferring products between stages. The outgoing material from stage $i$ serves as the feed material for production at stage $i-1$. Stages are numbered in ascending order: $0, 1, ..., M-1$, with stage 0 being the retailer. Production at each stage is limited by the stage's production capacity and available inventory. Figure \ref{fig:flowchart} depicts the flow of raw materials through various stages of production and inventory management, ultimately culminating in the fulfillment of customer demand at the retail level.

There are $T$ periods in each simulation, starting from 1, with $t = 0$ used for the initial condition of the supply chain. At the beginning of each time period, the following sequence of events occurs:
\begin{enumerate}
    \item Check deliveries: Each stage receives incoming inventory replenishment shipments that have arrived after the stage's respective lead time.
    \item Check orders and demands: Each stage places replenishment orders to their respective suppliers. Replenishment orders are filled according to the available production capacity and inventory at the suppliers. Customer demand occurs at the retailer and is filled based on the available  inventory at the retailer.
    \item Deliver orders and demands: Each stage delivers as many products as possible to satisfy  downstream demand or replenishment orders. Unfulfilled sales and replenishment orders are backlogged, with backlogged sales taking priority in the following period.
    \item Compute profits: Each stage computes the profit and cost for product sales, material orders, backlog penalties,
    and surplus inventory holding costs.
\end{enumerate}

\begin{table}[!h]
    \begin{tabular}{p{0.18\linewidth} p{0.70\linewidth}}
    \toprule
    \textbf{Notation} & \textbf{Definition} \\
    \midrule
    $m$ & Stage $\mathcal{M} = \{0, 1, 2, \ldots, M-1 \}$ \\
    $t$ & Period $\mathcal{T} = \{0, 1, 2, \ldots, T \}$ \\
    \midrule
    $I_{m,t}$ & Inventory at the end of period $t$ \\
    $\hat{I}_{m,t}$ & Desired inventory at the end of period $t$ \\
    $O_{m,t}$ & Requested order placed during period $t$ \\
    $R_{m,t}$ & Fulfilled order during period $t$ \\
    $D_t$ & Customer demand during period $t$ \\
    $S_{m,t}$ & Sales during period $t$ \\
    $B_{m,t}$ & Backlog at the end of period $t$ \\
    $L_m$ & Lead times between stage $m+1$ and stage $m$ \\
    $L_{\max}$ & Maximum lead time in the system \\
    $P_{m,t}$ & Profit at stage $m$ during period $t$ \\
    \midrule
    $c_m$ & Production capacity at stage $m$ \\
    $p_m$ & Unit sale price \\
    $r_m$ & Unit order (procurement) cost \\
    $k_m$ & Unit penalty for unfulfilled order \\
    $h_m$ & Unit inventory holding cost \\
    \bottomrule
    \end{tabular}
    \caption{Notations and definitions for parameters.}
    \label{tab:notation}
\end{table}

With the notations defined in Table \ref{tab:notation}, the entire inventory management problem (IMP), inspired by \citet{hubbs2020or}, can be expressed using following equations:
\begin{align}
    & I_{m,t} = I_{m,t-1} + R_{m,t-L_m} - S_{m,t}, \quad \forall m \in \mathcal{M}, \label{eq:1} \tag{1} \\
    & R_{m,t} = \min (B_{m+1,t-1} + O_{m,t}, c_{m+1}, I_{m+1,t-1} + \notag \\
    & \quad R_{m+1,t-L_{m+1}}), \quad \forall m = 0, ..., M - 2, \label{eq:2a} \tag{2a} \\
    & R_{M-1,t} = O_{M-1,t},  \label{eq:2b} \tag{2b} \\
    & S_{m,t} = R_{m-1,t}, \quad \forall m = 1, ..., M - 1, \label{eq:3a} \tag{3a} \\
    & S_{0,t} = \min (B_{0,t-1} + D_{t}, c_0, I_{0,t-1} + R_{0,t-L_0}), \label{eq:3b} \tag{3b} \\
    & B_{m,t} = B_{m,t-1} + O_{m-1,t} - S_{m,t}, \label{eq:4a} \tag{4a} \\
    & \quad \forall m = 1, ..., M - 1, \notag \\
    & B_{0,t} = B_{0,t-1} + D_{t} - S_{0,t}, \label{eq:4b} \tag{4b} \\
    &P_{m,t} = p_{m} S_{m,t} - r_{m} R_{m,t} - k_{m} B_{m,t} - h_m I_{m,t}, \label{eq:5} \tag{5} \\ 
    & \quad \forall m \in \mathcal{M} . \notag
\end{align}

In Equation \ref{eq:1}, the current inventory at stage $m$ at the end of the current period $t$ is equal to the final inventory in the previous period, plus fulfilled order placed $L_m$ periods ago, minus the sales during the current period. In Equation \ref{eq:2a}, fulfilled order at stage $m$ placed during period $t$ is decided by (1) previous backlog at the upstream stage plus newly requested orders, (2) upstream stage production capacity, and (3) total available inventory at the upstream stage $m+1$ at the start of period $t$, including leftover stock from the previous period and newly arrived orders after accounting for lead time. The final fulfilled order is the minimum of these three conditions, ensuring that the order does not exceed any of these constraints. Equation \ref{eq:2b} tells us requested orders at the upmost stage are always fulfilled, because we assume an unlimited supply of raw materials. Sales are always equal to fulfilled orders except at stage 0 (retailer), as shown in Equation \ref{eq:3a}. In Equation \ref{eq:3b}, sales at stage 0 (retailer) during period $t$ are determined by the minimum of three conditions: (1) the previous backlog at stage 0 plus the current customer demand, (2) the production capacity of stage 0, and (3) the total available inventory at stage 0 at the start of period $t$, which includes leftover stock from the previous period and newly fulfilled orders after accounting for the lead time. In Equation \ref{eq:4a}, the backlog at stage $m$ during period $t$ is calculated as the sum of the previous period's backlog at stage and the orders requested from the previous stage, minus the sales at stage $m$, for all stages except the retailer. In Equation \ref{eq:4b}, the backlog at stage 0 (retailer) during period $t$ is calculated similarly to Equation \ref{eq:4a}, but the requested order is replaced by customer demand because the retailer is directly in contact with customers. In Equation \ref{eq:5}, the profit at each stage $m$ during period $t$ is calculated as the sales revenue minus the procurement costs, unfulfilled order costs, and inventory holding costs.

\subsection{InvAgent Model}
In this work, we propose InvAgent, a LLM based multi-agent inventory management system for supply chain optimization. InvAgent includes several key agents: one user proxy and one agent for each stage. The user proxy serves as an intermediary between the environment and all supply chain agents, facilitating communication and managing the exchange of data. The framework of InvAgent method is illustrated in Figure \ref{fig:framework}, which follows these steps:
\begin{enumerate}
    \item The user proxy resets the environment at the beginning of the first round.
    \item The user proxy requests the state of the current round for each stage from the environment. \label{step:state}
    \item The user proxy provides the state to each stage and requests the action from it.
    \item The user proxy sends the agent actions to the environment and obtains the next state and the reward for this step.
    \item The user proxy determines whether the simulation is terminated; if not, the simulation moves to step \ref{step:state}.
\end{enumerate}
At the beginning of the simulation, we create system messages for agents in Figure \ref{fig:system}, which provide essential information, such as definitions, roles, and goals in the supply chain. The state $s_{m,t}$ and action $a_{m,t}$ of an agent are defined as: $s_{m,t} = [c_m, p_m, r_m, k_m, h_m, L_m, I_{m,t-1}, B_{m,t-1}, \\ B_{m+1,t-1}, S_{m,t-L_{\max}}, \dots, S_{m,t-1}, 0, \dots, 0, \\ R_{m,t-L_m}, \dots, R_{m,t-1}]$ and $a_{m,t} = O_{m,t}$, where the state includes the current stage features, inventory, backlog, upstream backlog, recent sales, and arriving deliveries with left zero padding.

\begin{figure}[!h]
  \centering
  \begin{tcolorbox}
    \textbf{System Message:}\\
    \textbf{Retailer:} You play a crucial role in a 4-stage supply chain as the stage 1 (retailer). Your goal is to minimize the total cost by managing inventory and orders effectively.\\
    \textbf{Wholesaler:} You play a crucial role in a 4-stage supply chain as the stage 2 (wholesaler). Your goal is to minimize the total cost by managing inventory and orders effectively.\\
    \textbf{Distributor:} You play a crucial role in a 4-stage supply chain as the stage 3 (distributor). Your goal is to minimize the total cost by managing inventory and orders effectively.\\
    \textbf{Manufacturer:} You play a crucial role in a 4-stage supply chain as the stage 4 (manufacturer). Your goal is to minimize the total cost by managing inventory and orders effectively.
  \end{tcolorbox}
  \caption{System messages providing essential information, such as definitions, roles, and goals in the supply chain.}
  \label{fig:system}
\end{figure}

The prompt, as designed in Figure \ref{fig:prompt}, aims to provide the state and request actions from each agent, ensuring effective decision-making and clear communication within the supply chain. It includes contextual information such as the current period, stage\footnote{The stage is counted from 1 instead of 0 in the prompt to prevent confusion for LLMs.}, and number of stages to position the model within the supply chain. The state description (Figure \ref{fig:state}) provides a comprehensive snapshot of inventory levels, backlogs, previous sales, and incoming deliveries, enabling informed decisions. Demand (Figure \ref{fig:demand}) and downstream order (Figure \ref{fig:downstream}) details help match supply with immediate needs, allowing upstream suppliers to quickly respond to downstream orders or demands. The strategy description (Figure \ref{fig:strategy}) outlines guidelines like considering lead times and avoiding overordering to maintain inventory balance. By requesting reasoning before specifying the action, the prompt promotes transparency and interpretability in decision-making. This design leverages LLMs' capabilities to enhance inventory management, ensuring decisions are well-informed, transparent, and aligned with the supply chain strategy. One example of the prompt and the response from GPT-4 is shown in Section \ref{sec:example}.

\begin{figure}[!h]
  \centering
  \begin{tcolorbox}
    \textbf{Prompt:}
    
    Now this is the round \{Period\}, and you are at the stage \{Stage\} of \{Number of  Stages\} in the supply chain. Given your current state:\\
    \{State Description\}\\
    
    \{Demand Description\} \{Downstream Order Description\} What is your action (order quantity) for this round?\\

    \{Strategy Description\}\\

    Please state your reason in 1-2 sentences first and then provide your action as a non-negative integer within brackets (e.g. [0]).
  \end{tcolorbox}
  \caption{Prompt provided to LLMs for inventory management simulation. State description, demand description, downstream order description, and strategy description are shown in Figures \ref{fig:state}, \ref{fig:demand}, \ref{fig:downstream}, and \ref{fig:strategy}, respectively.}
  \label{fig:prompt}
\end{figure}

\begin{figure}[!h]
  \centering
  \begin{tcolorbox}
    \textbf{State Description:}\\
    \ - Lead Time: \{Lead Time\} round(s)\\
    \ - Inventory Level: \{Inventory\} unit(s)\\
    \ - Current Backlog (you owing to the downstream): \{Backlog\} unit(s)\\
    \ - Upstream Backlog (your upstream owing to you): {Upstream Backlog} unit(s)\\
    \ - Previous Sales (in the recent round(s), from old to new): \{Sales\}\\
    \ - Arriving Deliveries (in this and the next round(s), from near to far): \{Deliveries\}
  \end{tcolorbox}
  \caption{State descriptions providing the current state for each agent in each period. For the previous sales, we select recent $L_{max}$ periods, and for the arriving deliveries, we select next $L_m$ periods.}
  \label{fig:state}
\end{figure}

\begin{figure}[!h]
  \centering
  \begin{tcolorbox}
    \textbf{Demand Description:}\\
    \textbf{Constant Demand:} The expected demand at the retailer (stage 1) is a constant 4 units for all 12 rounds.\\
    \textbf{Variable Demand:} The expected demand at the retailer (stage 1) is a discrete uniform distribution U\{0, 4\} for all 12 rounds.\\
    \textbf{Larger Demand:} The expected demand at the retailer (stage 1) is a discrete uniform distribution U\{0, 8\} for all 12 rounds.\\
    \textbf{Seasonal Demand:} The expected demand at the retailer (stage 1) is a discrete uniform distribution U\{0, 4\} for the first 4 rounds, and a discrete uniform distribution U\{5, 8\} for the last 8 rounds.\\
    \textbf{Normal Demand:} The expected demand at the retailer (stage 1) is a normal distribution N(4, 2\string^2), truncated at 0, for all 12 rounds.
  \end{tcolorbox}
  \caption{Demand descriptions for different demand scenarios included in the LLM prompt.}
  \label{fig:demand}
\end{figure}

\begin{figure}[!h]
  \centering
  \begin{tcolorbox}
    \textbf{Downstream Order Description:}\\
    Your downstream order from the stage \{Stage - 1\} for this round is \{Actions[Stage - 1]\}.
  \end{tcolorbox}
  \caption{The downstream order from the previous stage to the current stage at one round, which can delivery the downstream information faster.}
  \label{fig:downstream}
\end{figure}

\begin{figure}[!h]
  \centering
  \begin{tcolorbox}
    \textbf{Strategy Description:}
    
    Golden rule of this game: Open orders should always equal to "expected downstream orders + backlog". If open orders are larger than this, the inventory will rise (once the open orders arrive). If open orders are smaller than this, the backlog will not go down and it may even rise. Please consider the lead time and place your order in advance. Remember that your upstream has its own lead time, so do not wait until your inventory runs out. Also, avoid ordering too many units at once. Try to spread your orders over multiple rounds to prevent the bullwhip effect. Anticipate future demand changes and adjust your orders accordingly to maintain a stable inventory level.
  \end{tcolorbox}
  \caption{Strategy description introducing the golden rule of the problem and providing the LLM with suggestions for decision-making.}
  \label{fig:strategy}
\end{figure}

The features of the prompt design are as follows: 1) \textbf{Zero-Shot Learning:} Our prompt operates on a zero-shot basis, requiring the LLM to generate responses based solely on its pre-existing knowledge and the information presented in the prompt. 2) \textbf{Demand Description:} Providing a clear and detailed description of the demand is crucial to ensure accurate understanding and effective responses, as there is no prior training process involved. 3) \textbf{Downstream Order:} The prompt considers downstream order, allowing swift information delivery and efficient sharing between different stages. 4) \textbf{Human-Crafted Strategy:} While the inherent strategy of the LLM is generally sufficient for simple scenarios, additional human-crafted strategies are assumed to enhance decision-making in more complex scenarios like seasonal demands. 5) \textbf{Chain-of-Thought (CoT):} The CoT approach enhances the explainability of results by guiding the LLM through a structured reasoning process, improving its understanding and reasoning capabilities, and leading to more accurate and reliable outcomes.

\subsection{Prompt and Response Example}
\label{sec:example}

An example of the InvAgent prompt and response for the constant demand scenario with GPT-4 is presented in Figure \ref{fig:example}. The prompt includes a detailed state description, a demand description specifying the expected demand at the retailer, and a strategy description advising aligning open orders with expected downstream orders and backlog after considering lead times and the bullwhip effect. In the response, the Retailer Agent, considering the current inventory suffices for up to 3 rounds of maximal demand and the 2-round lead time, decides not to place an order this round, aiming to prevent excessive inventory, as articulated in the agent's reasoned response.

\section{Experiments}

In this section, we evaluate the performance of InvAgent by describing experiment scenarios, baseline, settings, and results.

\subsection{Experiment Scenarios}

\begin{table*}[!h]
    \centering
    \begin{tabular}{llllll}
    \toprule
    \textbf{Scenario} & \textbf{Constant} & \textbf{Variable} & \textbf{Larger} & \textbf{Seasonal} & \textbf{Normal} \\
    \midrule
    Number of Stages & 4 & 4 & 4 & 4 & 4 \\
    Number of Periods & 12 & 12 & 12 & 12 & 12 \\
    Initial Inventories & [12, 12, 12, 12] & [12, 12, 12, 12] & [12, 12, 12, 12] & [12, 12, 12, 12] & [12, 14, 16, 18] \\
    Lead Times & [2,2,2,2] & [2, 2, 2, 2] & [2, 2, 2, 2] & [2, 2, 2, 2] & [1, 2, 3, 4] \\
    Demand & 4 & $\mathcal{U}(0, 4)$ & $\mathcal{U}(0, 8)$ & $\mathcal{U}(0, 4) \xrightarrow{5} \mathcal{U}(5, 8)$ & $\mathcal{N}(4,2^2)$ \\
    Product Capacities & [20, 20, 20, 20] & [20, 20, 20, 20] & [20, 20, 20, 20] &  [20, 20, 20, 20] & [20, 22, 24, 26] \\
    Sales Prices & [0, 0, 0, 0] & [0, 0, 0, 0] & [5, 5, 5, 5] &  [5, 5, 5, 5] & [9, 8, 7, 6] \\
    Order Costs & [0, 0, 0, 0] & [0, 0, 0, 0] & [5, 5, 5, 5] &  [5, 5, 5, 5] & [8, 7, 6, 5] \\
    Backlog Costs & [1, 1, 1, 1] & [1, 1, 1, 1] & [1, 1, 1, 1] & [1, 1, 1, 1] & [1, 1, 1, 1] \\
    Holding Costs & [1, 1, 1, 1] & [1, 1, 1, 1] & [1, 1, 1, 1] & [1, 1, 1, 1] & [1, 1, 1, 1] \\
    \bottomrule
    \end{tabular}
    \caption{Parameter settings for different supply chain scenarios ($\mathcal{U}$: discrete uniform distribution, $\mathcal{N}$: normal distribution, $\xrightarrow{n}$: a jump at the round $n$).}
    \label{tab:para}
\end{table*}

We design various experiment scenarios to evaluate the performance of our inventory management system in a multi-echelon supply chain, summarized in Table \ref{tab:para}. The scenarios range from a four-stage supply chain with constant demand to scenarios with increasing demand variability, seasonal patterns, and normally distributed demand. Each scenario introduces specific conditions, such as fluctuating demand, financial impacts, and varying operational constraints, to rigorously test the robustness and adaptability of the proposed model. These comprehensive scenarios provide a thorough test bed for assessing the efficacy and adaptability of our multi-agent system in managing dynamic inventory across a multi-echelon supply chain. More details about these scenarios are discussed in Appendix \ref{sec:scenarios}.

\subsection{Experiment Baselines}

We have four baselines for our experiments: two heuristic policies (base-stock policy and tracking demand policy) and two reinforcement learning (RL) policies (independent proximal policy optimization (IPPO) and multi-agent proximal policy optimization (MAPPO)). The heuristic baselines are designed to maintain sufficient inventory levels to fulfill customer demands or downstream orders. The stage order (action) is computed by: $O_{m,t} = \min(\max(0, \hat{O}_{m,t})$,
where $\hat{O}_{m,t}$ is determined by the desired inventory $\hat{I}_{m,t}$, current inventory $I_{m,t-1}$, upstream backlog order $B_{m+1,t-1}$, and cumulative sum of arriving deliveries: $\hat{O}_{m,t} = \hat{I}_{m,t} - I_{m,t-1} - B_{m+1,t-1} - \sum_{\Delta t = 1}^{L_{m}} R_{m,t - \Delta t} $.
For the \textbf{base-stock policy}, the desired inventory level is set equal to the production capacity:
\begin{equation}
    \hat{I}_{m,t} = c_m \label{eq:8} \tag{8}.
\end{equation}
For the \textbf{tracking demand policy}, the desired inventory is set as:
\begin{equation}
    \hat{I}_{m,t} = \bar{S}_{m,t-1} L_m + B_{m,t-1} \label{eq:9} \tag{9} ,
\end{equation}
with the average sale for recent rounds given by:
\begin{equation}
    \bar{S}_{m,t-1} = \frac{1}{L_{\max}} \sum_{\Delta t = 1}^{L_{\max}} S_{m,t - \Delta t} \label{eq:10} \tag{10} .
\end{equation}

\textbf{Independent PPO (IPPO)} updates policies by iteratively sampling data and optimizing a clipped surrogate objective function, with parameter sharing to improve learning efficiency. \textbf{Multi-agent PPO (MAPPO)} extends PPO for multi-agent environments, using a centralized value function to improve variance reduction and stability during training. 

More details about these baselines are discussed in Appendix \ref{sec:baselines}.

\subsection{Experiment Settings}

The performance of our model InvAgent is evaluated using the total reward from all stages all periods during one simulation (episode), and the reported numbers are averaged over 5 episodes for each experiment to reduce the uncertainty. We utilize the Python packages such as AutoGen \cite{wu2023autogen}, Gymnasium \cite{towers_gymnasium_2023}, and RLlib \cite{liang2018rllib}, and also LLMs including GPT-4, GPT-4O, and GPT-4-Turbo \cite{achiam2023gpt}. For the constant demand scenario, we change the last part of the prompt in Figure \ref{fig:prompt} to "([0], [4], or [8] only)" to boost the InvAgent performance. Additionally, the performance of baseline models is evaluated based on the episode reward averaged over 100 episodes. For reinforcement learning (RL), we explore various hyper-parameter settings in Appendix \ref{sec:hyperparameters}. The RL experiments are conducted on a NVIDIA A10 GPU.

\subsection{Experiment Results}

\begin{table*}[!h]
    \centering
    \begin{tabular}{llllll}
    \toprule
    \textbf{Model} & \textbf{Constant} & \textbf{Variable} & \textbf{Larger} & \textbf{Seasonal} & \textbf{Normal} \\
    \midrule
    Base-Stock & -296.00 (0.00) & -523.69 (49.15) & -392.21 (111.79) & -274.29 (40.75) & -322.44 (99.59) \\
    Tracking-Demand & -360.00 (0.00) & -412.41 (41.76) & -265.07 (99.67) & -421.90 (55.18) & -232.20 (75.45) \\
    IPPO & -132.17 (40.17) & -389.55 (40.28) & -202.39 (92.96) & -126.73 (183.63) & -102.90 (64.68) \\
    MAPPO & \textbf{-129.81} (16.02) & -391.53 (34.09) & \textbf{-106.79} (109.86) & \textbf{-99.39} (126.09) & \textbf{-41.98} (75.22) \\
    InvAgent (w/o strategy) & -156.00 (0.00) & \textbf{-336.60} (43.24) & -350.20 (149.57) & -488.00 (114.82) & -172.60 (104.70) \\
    InvAgent (w/ strategy) & -200.00 (0.00) & -377.60 (53.50) & -357.60 (50.04) & -420.60 (225.42) & -192.40 (98.51) \\
    \bottomrule
    \end{tabular}
    \caption{Mean episode rewards and standard deviations (in parentheses) for base-stock, tracking-demand, IPPO, MAPPO, and InvAgent (with and without the hand-crafted strategy) models under various demand scenarios.}
    \label{tab:results}
\end{table*}

\begin{table*}[!h]
    \centering
    \begin{tabular}{lllllllr}
    \toprule
    \textbf{Model} & \textbf{Demand} & \textbf{Downstream} & \textbf{Strategy}& \textbf{CoT} & \textbf{History} & \textbf{Reward} & \textbf{$\bm{\Delta}$\%} \\
    \midrule
    GPT-4 & \cmark & \cmark & \xmark & \cmark & \cmark & -336.60 (43.24) & 0.00\%  \\
    GPT-4 & \cmark & \cmark & \cmark & \cmark & \cmark & -377.60 (53.50) & -12.18\% \\
    GPT-4 & \xmark & \cmark & \cmark & \cmark & \cmark & -349.40 (29.43) & -3.80\% \\
    GPT-4 & \cmark & \xmark & \cmark & \cmark & \cmark & -419.00 (35.91) & -24.48\% \\
    GPT-4 & \xmark & \xmark & \cmark & \cmark & \cmark & -379.40 (40.03) & -12.72\% \\
    GPT-4 & \cmark & \cmark & \xmark & \cmark & \xmark & -339.20 (46.63) & -0.77\% \\
    GPT-4 & \cmark & \cmark & \cmark & \xmark & \cmark & -369.80 (36.83) & -9.86\% \\
    GPT-4 & \cmark & \cmark & \cmark & \cmark & \xmark & -387.40 (11.09) & -15.09\% \\
    GPT-4O & \cmark & \cmark & \cmark & \cmark & \cmark & -405.00 (35.14) & -20.32\% \\
    GPT-4-Turbo & \cmark & \cmark & \cmark & \cmark & \cmark & -636.40 (195.26) & -89.07\% \\
    \bottomrule
    \end{tabular}
    \caption{Ablation studies on different prompt settings of the InvAgent for the variable demand scenario, where each reward is averaged from 5 experiments and the standard deviation is reported in parentheses. The percentage change in the reward compared to the first result is also included.}
    \label{tab:ablation}
\end{table*}

Our experiment results, as shown in Table \ref{tab:results}, highlight the performance of various models across different demand scenarios. The InvAgent model demonstrates competitive performance, particularly in the variable demand scenario, where InvAgent (without the hand-crafted strategy) achieves the highest mean episode rewards. While the MAPPO model exhibits the best performance in the other demand scenarios, InvAgent’s zero-shot capability and adaptability offer significant benefits. This adaptability allows InvAgent to make reasonable decisions and understand concepts without specific examples, showcasing a level of generalization and adaptability akin to human intuition.

Compared to heuristic baselines, InvAgent demonstrates significant advantages by dynamically adapting to real-time conditions, thereby minimizing inventory costs and avoiding stockouts, particularly in variable demand scenarios, showcasing its effectiveness in managing unpredictable demand patterns. While RL models like MAPPO and IPPO achieve higher metrics in some scenarios due to extensive training, they also come with increased complexity, potential instability, and notable computational demands. In contrast, InvAgent's strengths lie in its explainability, ease of implementation, stability, and reasonable decision-making without prior training, making it a valuable alternative for dynamic inventory management despite not always outperforming RL models. The comparison of the InvAgent model with and without strategy shows that the inherent strategy of the LLM is generally sufficient for simpler scenarios, such as constant and variable demands. However, in more complex scenarios like seasonal demands, the addition of a human-crafted strategy enhances decision-making, proving beneficial for managing patterned demand and improving performance and adaptability.

\subsection{Ablation Studies}
\label{sec:ablation}

We conduct ablation studies to evaluate the impact of various prompt components in InvAgent under the variable demand scenario, as detailed in Table \ref{tab:ablation}. This scenario introduces randomness with demand varying uniformly between 0 and 4 units per period over 12 periods. The results demonstrate that the contributions of different components to the model's performance vary significantly, with the prompt without the strategy component performing the best. The ablation study confirms the robustness and adaptability of our model in dynamic supply chain environments. Components such as the demand description and downstream order are particularly essential for optimizing performance under variable demands. By keeping agent histories, all previous messages in the entire episode (simulation) are retained, allowing stage agents to use the entire chat history for context in their decision-making. Additionally, structured reasoning through chain-of-thought (CoT) also plays crucial roles. These findings emphasize the importance of each component in achieving effective and efficient inventory management, guiding future improvements and applications in more complex scenarios.

\section{Conclusion}

In this study, we demonstrate the effectiveness of using large language models (LLMs) as autonomous agents for multi-agent inventory management in supply chain optimization. Our novel model, InvAgent, leverages the zero-shot learning capabilities of LLMs, enabling adaptive and informed decisions without prior training. The integration of structured reasoning through the chain-of-thought methodology enhances the model's explainability and transparency, making it more reliable and easier to trust compared to traditional heuristic or reinforcement learning models. Experimental results show that our model performs competitively, achieving lower costs and better adaptability across various demand scenarios, highlighting the potential of LLMs to improve supply chain management by reducing inventory costs and minimizing stockouts.

In the future, the InvAgent model can be fine-tuned using reinforcement learning to enhance decision-making capabilities, allowing the LLMs to iteratively learn and optimize strategies. Additionally, real-world data can be used to evaluate the efficiency of InvAgent and the utility of its agents. For data exhibiting seasonality, decomposition into level, trend, and seasonal components can be explored to further refine predictive accuracy. Another key focus will be combining human-crafted strategies with the LLMs' capabilities to handle diverse and unpredictable demand patterns more robustly.

\bibliography{aaai25}

\appendix

\section{Prompt and Response Example}
\label{sec:example}

An example of the InvAgent prompt and response for the constant demand scenario with GPT-4 is presented in Figure \ref{fig:example}. The prompt includes a detailed state description, a demand description specifying the expected demand at the retailer, and a strategy description advising aligning open orders with expected downstream orders and backlog after considering lead times and the bullwhip effect. In the response, the Retailer Agent, considering the current inventory suffices for up to 3 rounds of maximal demand and the 2-round lead time, decides not to place an order this round, aiming to prevent excessive inventory, as articulated in the agent's reasoned response.

\begin{figure*}[!h]
  \centering
  \begin{tcolorbox}
    \textbf{Example:}

    \textbf{User Proxy (to Retailer Agent):}\\

    Now this is the round 1, and you are at the stage 1 of 4 in the supply chain. Given your current state:\\
    \ - Lead Time: 2 round(s)\\
    \ - Inventory Level: 12 unit(s)\\
    \ - Current Backlog (you owing to the downstream): 0 unit(s)\\
    \ - Upstream Backlog (your upstream owing to you): 0 unit(s)\\
    \ - Previous Sales (in the recent round(s), from old to new): [0, 0]\\
    \ - Arriving Deliveries (in this and the next round(s), from near to far): [0, 0]\\
    
    The expected demand at the retailer (stage 1) is a discrete uniform distribution $U\{0, 4\}$ for all 12 rounds. What is your action (order quantity) for this round?\\
    
    Golden rule of this game: Open orders should always equal to "expected downstream orders + backlog". If open orders are larger than this, the inventory will rise (once the open orders arrive). If open orders are smaller than this, the backlog will not go down and it may even rise. Please consider the lead time and place your order in advance. Remember that your upstream has its own lead time, so do not wait until your inventory runs out. Also, avoid ordering too many units at once. Try to spread your orders over multiple rounds to prevent the bullwhip effect. Anticipate future demand changes and adjust your orders accordingly to maintain a stable inventory level.\\
    
    Please state your reason in 1-2 sentences first and then provide your action as a non-negative integer within brackets (e.g. [0]).\\
    
    --------------------------------------------------------------------------------\\
    \textbf{Retailer Agent (to User Proxy):}\\
    
    Reason: Anticipating the future demand in the coming 2 rounds (considering the lead time) and keeping in mind that the current inventory level is enough for a maximal demand of 3 rounds, I think it's wise not to place an order this round to avoid excessive inventory.\\
    
    Action: [0]
  \end{tcolorbox}
  \caption{Example of the prompt and response of InvAgent for the constant demand scenario with GPT-4.}
  \label{fig:example}
\end{figure*}

\section{Experiment Details}

\subsection{More Details about Experiment Scenarios}
\label{sec:scenarios}

We describe the various experiment scenarios designed to evaluate the performance of our inventory management system in a multi-echelon supply chain. Each scenario introduces specific conditions to rigorously test the robustness and adaptability of the proposed model. Parameter settings for these scenarios are summarized in Table \ref{tab:para}.

In the first scenario, a four-stage supply chain is tested with a constant demand of 4 units per period over 12 periods, starting with 12 units of inventory per stage and a lead time of 2 periods. This scenario aims to test the basic functionality of the model under stable conditions. The second scenario introduces variable demand, uniformly ranging between 0 and 4 units per period, adding randomness to evaluate the system’s ability to manage fluctuating demand while maintaining efficient inventory. The third scenario further increases demand variability, with uniform distribution between 0 and 8 units per period, and incorporates sales and order costs set at 5 units per period, testing the model's capability to handle high variability and financial impacts. The fourth scenario simulates seasonal demand with a leaping pattern ranging from 4 to 8 units per period, maintaining the same financial parameters as the third scenario, to evaluate the system's performance under predictable but varying demand patterns. Finally, the fifth scenario features normally distributed demand with a mean of 4 units and a standard deviation of 2 units per period, varying lead times, initial inventories, sales prices and order costs across the stages. This scenario tests the system's performance under more realistic demand fluctuations and varying operational constraints. These scenarios collectively provide a comprehensive test bed for evaluating the efficacy and adaptability of our multi-agent system in managing dynamic inventory across a multi-echelon supply chain.

\subsection{More Details about Baselines}
\label{sec:baselines}

We have four baselines: two heuristic policies, namely the base-stock policy and the tracking demand policy, and two reinforcement learning (RL) policies, specifically independent proximal policy optimization (IPPO) and multi-agent proximal policy optimization (MAPPO). 

Two heuristic baselines are designed based on the desired inventory levels, where each stage aims to maintain sufficient inventory to fulfill customer demands or downstream orders. The stage order (action) is computed by
\begin{equation}
    O_{m,t} = \min(\max(0, \hat{O}_{m,t}), c_{m}) \label{eq:6} \tag{6},
\end{equation}
where the $\hat{O}_{m,t}$ is determined by the desired inventory $\hat{I}_{m,t}$, current inventory $I_{m,t-1}$, upstream backlog order $B_{m+1,t-1}$, and cumulative sum of arriving deliveries as
\begin{align}
    \hat{O}_{m,t} &= \hat{I}_{m,t} - I_{m,t-1} - B_{m+1,t-1} \notag \\*
    & \quad - \sum_{\Delta t = 1}^{L_{m}} R_{m,t - \Delta t} \label{eq:7} \tag{7} .
\end{align}
When an order is placed, the stage replenishes its inventory to this target level by ordering the difference between the base-stock level and the current inventory position with upstream backlog orders and arriving deliveries. Based on different choices of the desired inventory, we introduce two heuristic polices as follows.

\textbf{Base-Stock Policy.} The base-stock policy \cite{lee1997information, oroojlooyjadid2022deep} is an inventory management strategy where each stage in the supply chain maintains a constant inventory level, or base-stock level. Here, the desired inventory level is set equal to the production capacity:
\begin{equation}
    \hat{I}_{m,t} = c_m \label{eq:8} \tag{8}.
\end{equation}

\textbf{Tracking Demand Policy.} The tracking demand policy is an inventory management strategy that adjusts orders based on observed demand (or sale) patterns rather than maintaining a constant base-stock level. By dynamically aligning supply with actual consumption, this policy ensures a responsive and efficient inventory system. In this policy, the desired inventory is set as:
\begin{equation}
    \hat{I}_{m,t} = \bar{S}_{m,t-1} L_m + B_{m,t-1} \label{eq:9} \tag{9} ,
\end{equation}
where the average sale for recent rounds is
\begin{equation}
    \bar{S}_{m,t-1} = \frac{1}{L_{\max}} \sum_{\Delta t = 1}^{L_{\max}} S_{m,t - \Delta t} \label{eq:10} \tag{10} .
\end{equation}
More heuristic baselines and their results are discussed in Appendix \ref{sec:variants}.

\textbf{Independent Proximal Policy Optimization (IPPO) with Parameter Sharing.} Proximal policy optimization (PPO) \cite{schulman2017proximal} updates policies by iteratively sampling data through interaction with the environment and optimizing a clipped surrogate objective function using multiple epochs of stochastic gradient ascent. Independent PPO (IPPO) \cite{de2020independent} is a RL approach where each agent is independently trained with the PPO algorithm. Here we employ the parameter sharing by using the same policy parameters for all agents to improve learning efficiency and coordination.

\textbf{Multi-Agent Proximal Policy Optimization (MAPPO).} Multi-agent PPO (MAPPO) \cite{yu2022surprising} is an extension of PPO algorithm designed for multi-agent environments. It utilizes a centralized value function that takes into account global information from all agents, improving variance reduction and stability during training. 

\subsection{Reinforcement Learning Baseline Settings}
\label{sec:hyperparameters}

For reinforcement learning (RL), we explore various hyper-parameter settings, including the numbers of hidden units ([128, 128] and [256, 256]), activation function (ReLU), learning rate (1e-4, 5e-4, 1e-3), training batch size (500, 1000, 2000), stochastic gradient descent (SGD) minibatch size (32, 64, 128), number of SGD iterations (5, 10, 20), number of training iterations (500, 800, 1000, 1500), and discount factor (1.0). We randomly select 20 combinations of these hyperparameters during one experiment and keep the best one of them. The final hyperparameter settings used for the independent proximal policy optimization (IPPO) with parameter sharing and multi-agent proximal policy optimization (MAPPO) baselines are provided in Table \ref{tab:IPPO_para} and Table \ref{tab:MAPPO_para} respectively.

\begin{table*}[!h]
    \centering
    \begin{tabular}{llllll}
    \toprule
    \textbf{Hyperparameter} & \textbf{Constant} & \textbf{Variable} & \textbf{Larger} & \textbf{Seasonal} & \textbf{Normal} \\
    \midrule
    Numbers of Hidden Unit & [128, 128] & [256, 256] & [128, 128] & [128, 128] & [128, 128] \\
    Activation Function & ReLU & ReLU & ReLU & ReLU & ReLU \\
    Learning Rate & 0.0001 & 0.0001 & 0.001 & 0.0005 & 0.0005 \\
    Training Batch Size & 1000 & 1000 & 2000 & 2000 & 1000 \\
    SGD Minibatch Size & 128 & 128 & 128 & 128 & 128 \\
    Number of SGD Iterations & 5 & 10 & 5 & 5 & 5 \\
    Number of Training Iterations & 1000 & 1500 & 1000 & 800 & 500 \\
    \bottomrule
    \end{tabular}
    \caption{Hyperparameters for the independent proximal policy optimization (IPPO) with parameter sharing baseline.}
    \label{tab:IPPO_para}
\end{table*}

\begin{table*}[!h]
    \centering
    \begin{tabular}{llllll}
    \toprule
    \textbf{Hyperparameter} & \textbf{Constant} & \textbf{Variable} & \textbf{Larger} & \textbf{Seasonal} & \textbf{Normal} \\
    \midrule
    Numbers of Hidden Unit & [128, 128] & [128, 128] & [128, 128] & [256, 256] & [128, 128] \\
    Activation Function & ReLU & ReLU & ReLU & ReLU & ReLU \\
    Learning Rate & 0.0001 & 0.0001 & 0.001 & 0.0001 & 0.0001 \\
    Training Batch Size & 500 & 2000 & 2000 & 1000 & 500 \\
    SGD Minibatch Size & 128 & 32 & 32 & 128 & 128 \\
    Number of SGD Iterations & 10 & 5 & 10 & 10 & 10 \\
    Number of Training Iterations & 500 & 500 & 800 & 1500 & 500 \\
    \bottomrule
    \end{tabular}
    \caption{Hyperparameters for the multi-agent proximal policy optimization (MAPPO) baseline.}
    \label{tab:MAPPO_para}
\end{table*}

\subsection{Evaluation Results of Heuristic Baseline Variants}
\label{sec:variants}

\begin{table*}[!h]
    \centering
    \begin{tabular}{llllll}
    \toprule
    \textbf{Desired Inventory} & \textbf{Constant} & \textbf{Variable} & \textbf{Larger} & \textbf{Seasonal} & \textbf{Normal} \\
    \midrule
    $0.8 c_m$ & -208.00 (0.00) & -435.69 (49.15) & \textbf{-234.28} (102.81) & \textbf{-207.75} (34.67) & \textbf{-150.67} (101.80) \\
    $0.9 c_m$ & -252.00 (0.00) & -479.69 (49.15) & -310.74 (109.16) & -229.08 (34.66) & -226.31 (103.32) \\
    $c_m$ & -296.00 (0.00) & -523.69 (49.15) & -392.21 (111.79) & -274.29 (40.75) & -322.44 (99.59) \\
    \midrule
    $S_{m,t-1} L_m + B_{m,t-1}$ & -364.00 (0.00) & -390.17 (44.24) & -393.31 (79.05) & -525.84 (47.85) & -283.39 (61.83) \\
    $S_{m,t-1} (L_m + 1) + B_{m,t-1}$ & \textbf{-120.00} (0.00) & -395.68 (41.86) & -470.55 (76.73) & -524.26 (64.68) & -351.23 (90.08) \\
    $\bar{S}_{m,t-1} L_m + B_{m,t-1}$ & -360.00 (0.00) & -412.41 (41.76) & -265.07 (99.67) & -421.90 (55.18) & -232.20 (75.45) \\
    $\bar{S}_{m,t-1} (L_m + 1) + B_{m,t-1}$ & -252.00 (0.00) & \textbf{-382.77} (48.50) & -489.75 (110.96) & -610.03 (94.43) & -177.54 (70.87) \\
    $1.2 \bar{S}_{m,t-1} L_m + B_{m,t-1}$ & -361.00 (0.00) & -397.22 (50.02) & -325.81 (98.39) & -479.07 (69.47) & -218.98 (73.26) \\
    \bottomrule
    \end{tabular}
    \caption{Evaluation results (averaged episode rewards and their standard deviations) for different heuristic model variants under various demand scenarios.}
    \label{tab:variants}
\end{table*}

The evaluation results of the base-stock policy and the tracking demand policy variants are displayed in Table \ref{tab:variants}. These evaluations provide insights into how different inventory policies perform under diverse demand scenarios. The first section of the table shows the performance of the base-stock policy with different desired inventory levels based on production capacities $c_m$. The results indicate that maintaining a lower desired inventory level generally results in better performance across varying demand conditions. The second section of the table includes five variants of the tracking demand policy, denoted by different formulas involving the sales $S_{m,t-1}$, lead time $L_m$, and backlog $B_{m,t-1}$. While no single variant consistently outperforms the others, averaging sales typically helps manage variable demands across most scenarios.

\section{Case Studies}

In this section, we have two demand scenario case studies based on our model InvAgent. One case study examines a variable demand scenario without strategy, while the other looks at a seasonal demand scenario with strategy in place. The supply chain in these scenarios comprises four stages, moving from downstream to upstream: the retailer, wholesaler, distributor, and manufacturer.

\subsection{Variable Demand Scenario}

\begin{figure*}[!h]
  \centering
  \includegraphics[width=.65\textwidth]{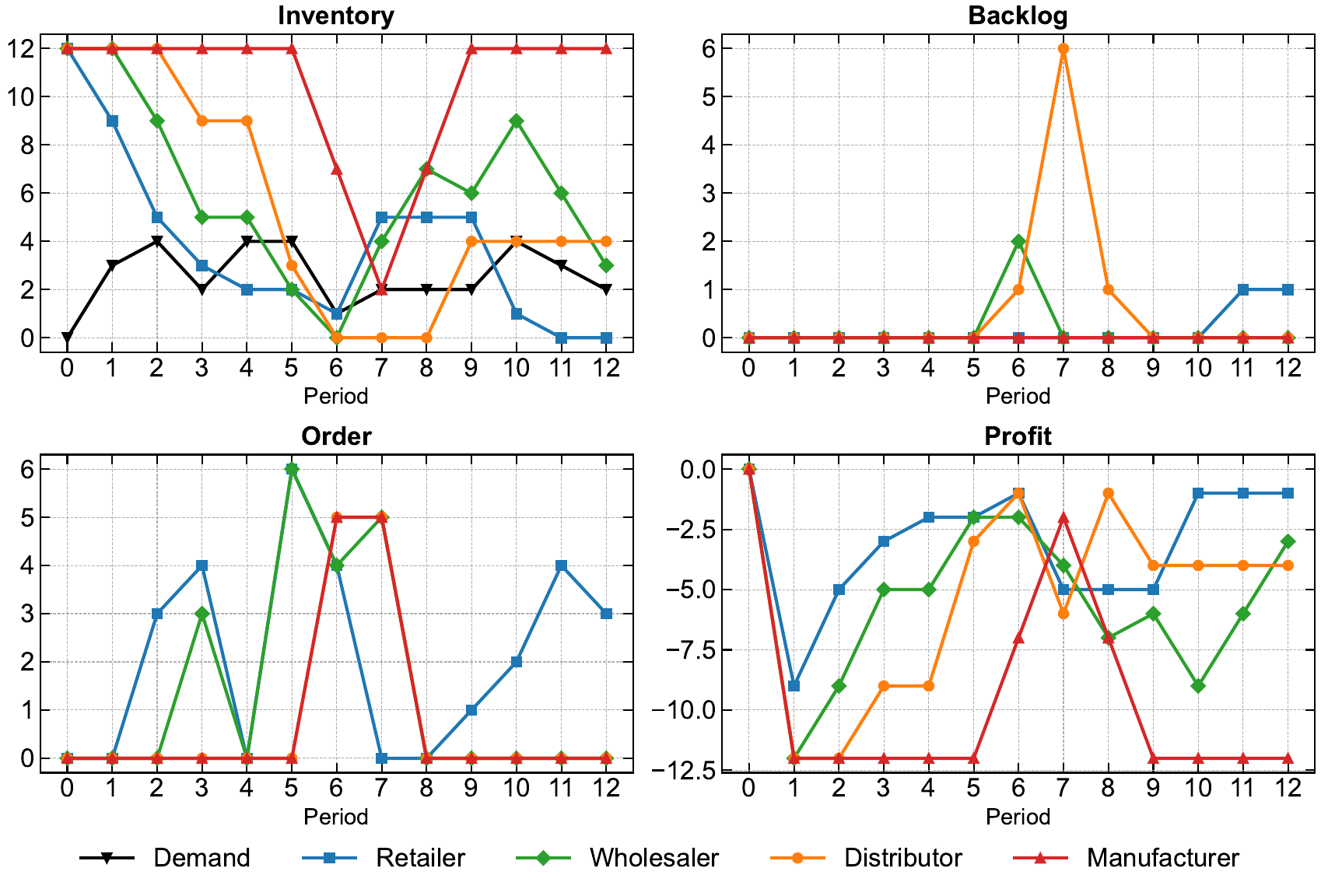}
  \caption{Inventory, backlog, orders, and profit analyses for the variable demand scenario in one entire episode (simulation) for retailer, wholesaler, distributor, and manufacturer agents in the InvAgent model without strategy.}
  \label{fig:variable_demand}
\end{figure*}

In the variable demand scenario, Figure \ref{fig:variable_demand} shows how LLMs take actions (orders) in response to changes in the demand, inventory, backlog, and profit. At the start of the simulation (episode), when demand first appears, the retailer begins to respond to the change. Initially, the retailer's inventory decreases because retailers are the first to supply customers, followed by the wholesaler, distributor, and manufacturer. Due to the lead time from the upstream suppliers, the retailer's inventory cannot stabilize immediately, even after placing orders with the upstream wholesaler. After several ordering cycles, the retailer's inventory eventually reaches a relatively steady state.

Another interesting phenomenon occurs in the middle of the simulation, when the backlog value of the distributor reaches its peak. This happens because the distributor's inventory starts decreasing several periods earlier and completely runs out in period 6. The distributor fails to restock in a timely manner as the inventory dwindles, resulting in a huge backlog in period 7. To prevent this, the distributor should place orders at least the lead time periods before the inventory running out.

\subsection{Seasonal Demand Scenario}

\begin{figure*}[!h]
  \centering
  \includegraphics[width=.65\textwidth]{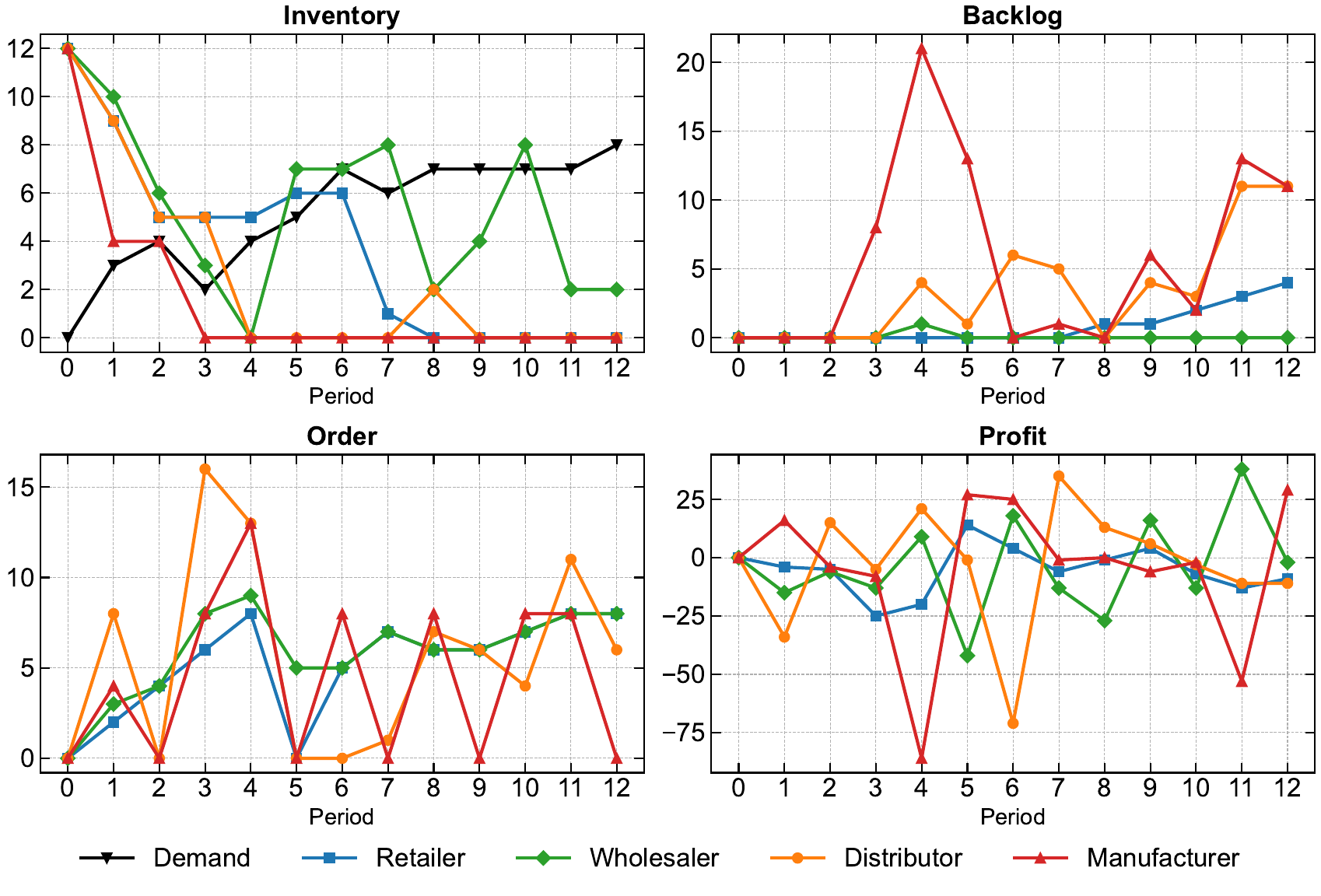}
  \caption{Inventory, backlog, orders, and profit analyses for the seasonal demand scenario in one entire episode (simulation) for retailer, wholesaler, distributor, and manufacturer agents in the InvAgent model with strategy.}
  \label{fig:seasonal_demand}
\end{figure*}

In the seasonal demand scenario, Figure \ref{fig:seasonal_demand} shows how LLM takes action (order) in response to changes in demand, inventory, backlog, and profit. In this scenario, all agents are informed of the demand distribution in each period. Specifically, the demand follows a uniform distribution, $\mathcal{U}(0, 4)$, from periods 1 to 4, and a different uniform distribution, $\mathcal{U}(5, 8)$, from periods 5 to 12. During periods 3 and 4, four agents, particularly the distributor and manufacturer, attempt to order large quantities of products from their upstream suppliers. The manufacturer's inventory is exhausted due to the high volume of downstream orders. This leads to a spike in backlog during period 4, causing the manufacturer's profit to reach its minimum value. In the subsequent periods, the manufacturer continues to order raw materials, which helps mitigate the backlog and demonstrates the flexibility and resilience of our model.

\end{document}